\documentclass[letterpaper, 10 pt, conference]{class/ieeeconf}  

\IEEEoverridecommandlockouts                              
\usepackage{float}
\usepackage{verbatim} 
\usepackage{graphicx}
\usepackage{amsmath}
\usepackage{amssymb}
\usepackage{latexsym}
\usepackage{url}
\usepackage{cite}
\usepackage{relsize}
\usepackage{multirow}
\usepackage{afterpage}
\usepackage{ifthen}
\usepackage{graphicx}
\usepackage{algpseudocode,algorithm,algorithmicx}
\usepackage{subfigure}
\usepackage{flushend}
\usepackage{epstopdf}
\usepackage{stackengine}
\usepackage{gensymb}
\usepackage[bookmarks=false, linkcolor=blue, urlcolor=blue, citecolor=blue]{hyperref} 
\usepackage{booktabs}
\usepackage{makecell}
\usepackage{hhline}
\usepackage{courier}
\usepackage{lipsum}
\usepackage[dvipsnames]{xcolor}
\usepackage{copyrightbox} 

\hypersetup{pdfstartview={FitH}}
\title{\LARGE \bf Tracking Everything in Robotic-Assisted Surgery}

\author{Bohan Zhan$^{1}$, Wang Zhao$^{2}$, Yi Fang$^{3}$, Bo Du$^{4}$, Francisco Vasconcelos$^{5}$, Danail Stoyanov$^{5}$, \\ Daniel S. Elson$^{1}$, Baoru Huang$^{1,5,6}$
\thanks{$^1$The Hamlyn Centre for Robotic Surgery, Imperial College London, SW7 2AZ, UK {\tt Baoru.Huang18@imperial.ac.uk}}
\thanks{$^2$Department of Computer Science, Tsinghua University, Beijing, China}
\thanks{$^3$Embodied AI and Robotics Lab, NYU Abu Dhabi, Abu Dhabi, UAE}
\thanks{$^4$School of Computer Science, Wuhan University, Wuhan, China}
\thanks{$^5$Hawkes Institute, University College London, WC1E 6BT, UK}
\thanks{$^6$Department of Computer Science, University of Liverpool, L69 7ZX, UK}
}

\begin{document}

\newtheorem{problem}{Problem}
\newtheorem{lemma}{Lemma}
\newtheorem{theorem}[lemma]{Theorem}
\newtheorem{claim}{Claim}
\newtheorem{corollary}[lemma]{Corollary}
\newtheorem{definition}[lemma]{Definition}
\newtheorem{proposition}[lemma]{Proposition}
\newtheorem{remark}[lemma]{Remark}
\newenvironment{LabeledProof}[1]{\noindent{\it Proof of #1: }}{\qed}

\def\beq#1\eeq{\begin{equation}#1\end{equation}}
\def\bea#1\eea{\begin{align}#1\end{align}}
\def\beg#1\eeg{\begin{gather}#1\end{gather}}
\def\beqs#1\eeqs{\begin{equation*}#1\end{equation*}}
\def\beas#1\eeas{\begin{align*}#1\end{align*}}
\def\begs#1\eegs{\begin{gather*}#1\end{gather*}}

\newcommand{\poly}{\mathrm{poly}}
\newcommand{\eps}{\epsilon}
\newcommand{\e}{\epsilon}
\newcommand{\polylog}{\mathrm{polylog}}
\newcommand{\rob}[1]{\left( #1 \right)} 
\newcommand{\sqb}[1]{\left[ #1 \right]} 
\newcommand{\cub}[1]{\left\{ #1 \right\} } 
\newcommand{\rb}[1]{\left( #1 \right)} 
\newcommand{\abs}[1]{\left| #1 \right|} 
\newcommand{\zo}{\{0, 1\}}
\newcommand{\zonzo}{\zo^n \to \zo}
\newcommand{\zokzo}{\zo^k \to \zo}
\newcommand{\zot}{\{0,1,2\}}
\newcommand{\en}[1]{\marginpar{\textbf{#1}}}
\newcommand{\efn}[1]{\footnote{\textbf{#1}}}
\newcommand{\vecbm}[1]{\boldmath{#1}} 
\newcommand{\uvec}[1]{\hat{\vec{#1}}}
\newcommand{\thv}{\vecbm{\theta}}
\newcommand{\junk}[1]{}
\newcommand{\var}{\mathop{\mathrm{var}}}
\newcommand{\rank}{\mathop{\mathrm{rank}}}
\newcommand{\diag}{\mathop{\mathrm{diag}}}
\newcommand{\tr}{\mathop{\mathrm{tr}}}
\newcommand{\acos}{\mathop{\mathrm{acos}}}
\newcommand{\atantwo}{\mathop{\mathrm{atan2}}}
\newcommand{\SVD}{\mathop{\mathrm{SVD}}}
\newcommand{\quadf}{\mathop{\mathrm{q}}}
\newcommand{\linterp}{\mathop{\mathrm{l}}}
\newcommand{\sgn}{\mathop{\mathrm{sign}}}
\newcommand{\sym}{\mathop{\mathrm{sym}}}
\newcommand{\avg}{\mathop{\mathrm{avg}}}
\newcommand{\mean}{\mathop{\mathrm{mean}}}
\newcommand{\erf}{\mathop{\mathrm{erf}}}
\newcommand{\grad}{\nabla}
\newcommand{\R}{\mathbb{R}}
\newcommand{\defeq}{\triangleq}
\newcommand{\dims}[2]{[#1\!\times\!#2]}
\newcommand{\sdims}[2]{\mathsmaller{#1\!\times\!#2}}
\newcommand{\udims}[3]{#1}
\newcommand{\udimst}[4]{#1}
\newcommand{\com}[1]{\rhd\text{\emph{#1}}}
\newcommand{\ind}{\hspace{1em}}
\newcommand{\argmin}[1]{\underset{#1}{\operatorname{argmin}}}
\newcommand{\floor}[1]{\left\lfloor{#1}\right\rfloor}
\newcommand{\step}[1]{\vspace{0.5em}\noindent{#1}}
\newcommand{\quat}[1]{\ensuremath{\mathring{\mathbf{#1}}}}
\newcommand{\norm}[1]{\left\lVert#1\right\rVert}
\newcommand{\ignore}[1]{}
\newcommand{\specialcell}[2][c]{\begin{tabular}[#1]{@{}c@{}}#2\end{tabular}}
\newcommand*\Let[2]{\State #1 $\gets$ #2}
\newcommand{\algorithmicbreak}{\textbf{break}}
\newcommand{\Break}{\State \algorithmicbreak}
\newcommand{\ra}[1]{\renewcommand{\arraystretch}{#1}}

\renewcommand{\vec}[1]{\mathbf{#1}} 

\algdef{S}[FOR]{ForEach}[1]{\algorithmicforeach\ #1\ \algorithmicdo}
\algnewcommand\algorithmicforeach{\textbf{for each}}
\algrenewcommand\algorithmicrequire{\textbf{Require:}}
\algrenewcommand\algorithmicensure{\textbf{Ensure:}}
\algnewcommand\algorithmicinput{\textbf{Input:}}
\algnewcommand\INPUT{\item[\algorithmicinput]}
\algnewcommand\algorithmicoutput{\textbf{Output:}}
\algnewcommand\OUTPUT{\item[\algorithmicoutput]}
\maketitle
\thispagestyle{empty}
\pagestyle{empty}

\begin{abstract}
Accurate tracking of tissues and instruments in videos is crucial for Robotic-Assisted Minimally Invasive Surgery (RAMIS), as it enables the robot to comprehend the surgical scene with precise locations and interactions of tissues and tools. Traditional keypoint-based sparse tracking is limited by featured points, while flow-based dense two-view matching suffers from long-term drifts. Recently, the Tracking Any Point (TAP) algorithm was proposed to overcome these limitations and achieve dense accurate long-term tracking. However, its efficacy in surgical scenarios remains untested, largely due to the lack of a comprehensive surgical tracking dataset for evaluation. To address this gap, we introduce a new annotated surgical tracking dataset for benchmarking tracking methods for surgical scenarios, comprising real-world surgical videos with complex tissue and instrument motions. We extensively evaluate state-of-the-art (SOTA) TAP-based algorithms on this dataset and reveal their limitations in challenging surgical scenarios, including fast instrument motion, severe occlusions, and motion blur, etc. 
Furthermore, we propose a new tracking method, namely SurgMotion, to solve the challenges and further improve the tracking performance. 
Our proposed method outperforms most TAP-based algorithms in surgical instruments tracking, and especially demonstrates significant improvements over baselines in challenging medical videos.
Our code and dataset are available at \textcolor{blue}{\href{https://github.com/zhanbh1019/SurgicalMotion}{https://github.com/zhanbh1019/SurgicalMotion}}.
\end{abstract}

\section{INTRODUCTION} \label{Sec:Intro}
Robotic-Assisted Minimally Invasive Surgery (RAMIS) is widely applied in various types of medical procedure~\cite{fiorini2022concepts}. Compared to traditional open surgery, RAMIS offers advantages such as reducing human errors, enhancing the capability for remote surgery, mitigating pain, and lowering the risk of infection \cite{diana2015robotic,wright2017robotic,zhang2018self}. The ultimate objective of RAMIS is fully automated surgery, which places high demands on the robot’s ability to understand the surgical scene \cite{yip2019robot,huang2022self}. Therefore, motion tracking plays a vital role in RAMIS. Precise tracking of tissues and surgical instruments enables the robot to locate instruments and target tissues, and assess their relative positions and interactions.

Traditional visual tracking algorithms in surgical scenarios have several limitations. Instrument tracking typically relies on box-based or segmentation-based methods, which struggle to capture detailed rotation and deformation of instruments effectively \cite{rueckert2024methods,huang2022simultaneous,cartucho2022enhanced}. Tissue tracking, on the other hand, often depends on sparse feature matching or dense optical flow methods \cite{schmidt2024tracking}. However, sparse feature matching performs poorly when handling deformable or weakly-textured tissues, and it only provides tracking on sparse feature points which may not fully fulfill the practical demands \cite{yip2012tissue}. While optical flow is a dense tracking method, it only tracks motion between adjacent frames, making it vulnerable to occlusion and prone to tracking drifts in medical videos \cite{ihler2020patient,huang2020tracking}.

\begin{figure}[t]
\centering
\includegraphics[width=1\linewidth,height=0.44\linewidth]{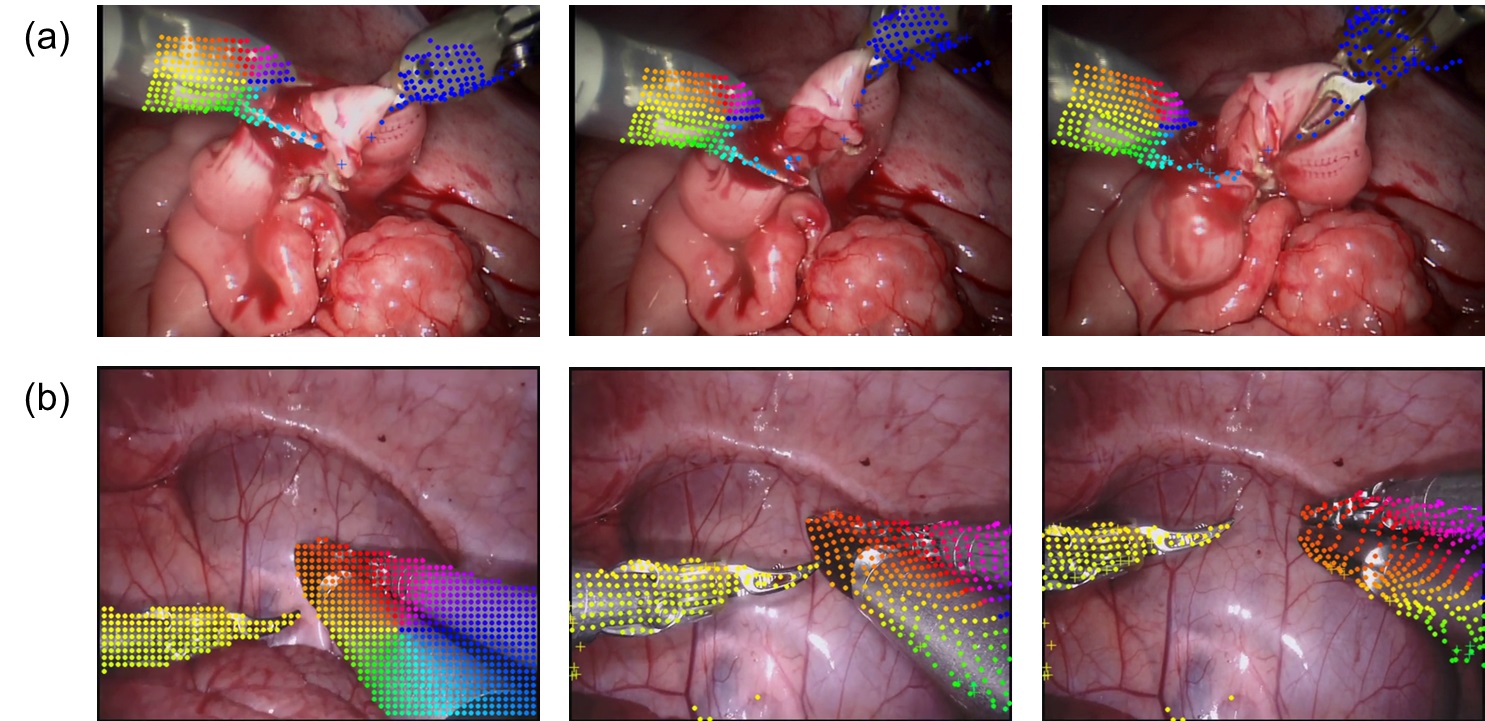}
\vspace{0.02cm}
\caption{The demonstration of our method for tracking every point across entire surgical video. (a), (b)  present results from different videos, with the tracking of instruments displayed from left to right over time.}
\label{fig:result}
\end{figure}

In the general field of motion tracking, similar challenges also exist. To address these limitations, recent research has proposed the TAP algorithm\cite{doersch2022tap}, which directly tracks every pixel across long-term videos and estimates occlusions via spatial-temporal feature learning and alignment, demonstrating superior performance over traditional tracking methods on a wide array of applications.
However, the TAP-based algorithm has not been employed in surgical environments. Most TAP-based methods are trained on synthetic datasets of natural scenes, which include precise point trajectories as ground truth \cite{ doersch2023tapir, harley2022particle, zheng2023pointodyssey, neoral2024mft, karaev2023cotracker}. The significant differences between natural scenes and surgical environments, such as poor lighting conditions, lack of significant texture features, and high specular reflection, present great challenges for applying TAP-based algorithms in surgical videos.

To evaluate the performance of TAP-based algorithms, several benchmarks have been established, including TAP-Vid \cite{doersch2022tap} and PointOdyssey \cite{zheng2023pointodyssey}, which feature manually or automatically annotated point trajectories. However, these benchmarks are limited to non-surgical domains, and there is a notable lack of equivalent datasets for surgical videos. In the context of surgical tracking, the SuPer dataset \cite{li2020super} and the SurgT dataset \cite{cartucho2024surgt} are the most relevant existing datasets.
Nevertheless, these datasets have significant limitations: the SuPer dataset has an insufficient number of labeled frames, while the SurgT dataset provides only sparse annotations, with a single labeled point per frame. As a result, existing datasets fall short in providing a comprehensive evaluation of TAP-based algorithms.

To address this gap, we developed a new dataset with manually labeled point trajectories to evaluate tracking algorithms in surgical scenarios. Specifically, we collected 20 real-world surgical videos, each with approximately 60 frames, and manually labeled 25 points per frame. Leveraging this dataset, we conducted a comprehensive evaluation of existing TAP-based algorithms, exposing their limitations in tracking fast-moving instruments in robotic surgery and other challenging regions. To further address these challenges, we designed an effective tracking method, SurgMotion, building upon OmniMotion~\cite{wang2023tracking} and proposed key constraints, including tool mask constraint, as-rigid-as-possible (ARAP) constraint, and sparse feature matching guidance. These innovations collectively yield significant improvements in tracking accuracy, particularly in challenging surgical videos. This method tracks any points, enabling simultaneous tracking of multiple surgical instruments without the need to distinguish between instrument categories, making it highly applicable to various surgical scenarios.
To summarize, the key contributions of this paper are as follows:
\begin{itemize}
    \item  We established a new surgical video tracking dataset, pioneering the exploration of tracking algorithms in real-world surgical scenarios.
    \item We benchmarked existing TAP-based algorithms on our dataset, revealing their limitations in tracking complex surgical motions in robotic surgery.
    \item We proposed a new tracking method, SurgMotion, for accurate tracking in surgical videos. Extensive experiments demonstrate that our method outperforms existing TAP-based methods and shows substantial improvements in challenging surgical videos.
\end{itemize}


\section{Related Work} \label{Sec:rw}
\subsection{Tracking Datasets in Surgical Domain}
Generating accurate tracking ground truth is highly challenging, and as a result, there is currently a lack of datasets for evaluating tracking algorithms in surgical scenarios. Existing datasets for tissue tracking, such as Semantic SuPer~\cite{lin2023semantic} and STIR~\cite{schmidt2024surgical}, use bead markers and indocyanine green to label real tissues. However, their precision is insufficient for accurately assessing single-point trajectories. The SuPer~\cite{li2020super} and SurgT~\cite{cartucho2024surgt} datasets, which are the closest to the requirements of TAP-based algorithms, offer manually annotated points as ground truth. However, the SuPer dataset contains only 52 frames with 20 points per frame, while SurgT, despite having 24,548 frames, provides just one annotated point per frame. Therefore, both the scale and the annotation accuracy of these datasets are inadequate for evaluating the performance of TAP-based algorithms.

\subsection{Surgical Scene Tracking}
Vision-based tracking in surgical scenarios is generally divided into tissue tracking and surgical instrument tracking. Tissue tracking relies on sparse feature matching or dense optical flow analysis. For example, a CNN-based optical flow method \cite{ihler2020patient} fine-tunes FlowNet~\cite{ilg2017flownet}, enabling a fast convolutional model for tissue tracking. The KINFlow~\cite{schmidt2022fast} uses k-nearest keypoint correspondences to track tissue motion. Additionally, SENDD \cite{schmidt2023sendd} employs Graph Neural Networks to match sparse keypoints and estimate per-point depth and 3D flow. Surgical instrument tracking is typically solved by box-based or segmentation-based algorithms. For example, YOLO-based tracking methods \cite{peng2022autonomous}, known for their real-time performance, locate instruments using bounding boxes. However, segmentation-based approaches offer higher accuracy. Transformer models like TraSeTR~\cite{zhao2022trasetr}, and MATIS~\cite{ayobi2023matis} demonstrate high precision in instrument tracking. Recently, foundation models, such as MedSAM~\cite{ma2024segment} and SurgicalSAM~\cite{yue2024surgicalsam}, further enhance the accuracy of surgical instrument segmentation and tracking.

\subsection{Tracking Any Point}
Tracking any point algorithms have been explored in recent years. PIPs \cite{harley2022particle} pioneers the concept of pixel tracking as a long-range motion estimation problem, enabling the tracking of every point in a video sequence within a small, fixed time window (8 frames). PIPs++ \cite{zheng2023pointodyssey} expands the temporal field of view of the original PIPs. Subsequently, TAP-Vid \cite{doersch2022tap} formalizes this problem and introduces a benchmark that includes manually annotated real-world datasets as well as a large set of synthetic datasets. TAP-Vid also proposes a simple baseline method, TAP-Net, which is combined with PIPs to create TAPIR \cite{doersch2023tapir}. MFT \cite{neoral2024mft} tracks every pixel in a template based on the combination of the optical flow field and different time spans. CoTracker \cite{karaev2023cotracker} presents a powerful transformer-based model to track points in the video by accounting for the correlation of points and tracking them jointly. Unlike the aforementioned methods trained on ground truth point trajectories from synthetic datasets, OmniMotion \cite{wang2023tracking} introduces a test-time optimization algorithm that requires no prior knowledge, achieving accurate, full-length motion estimation of every pixel in a video.

\section{Methodology} \label{Sec:method}
\begin{figure*}[h]
    \centering
    \includegraphics[width=0.8\linewidth]{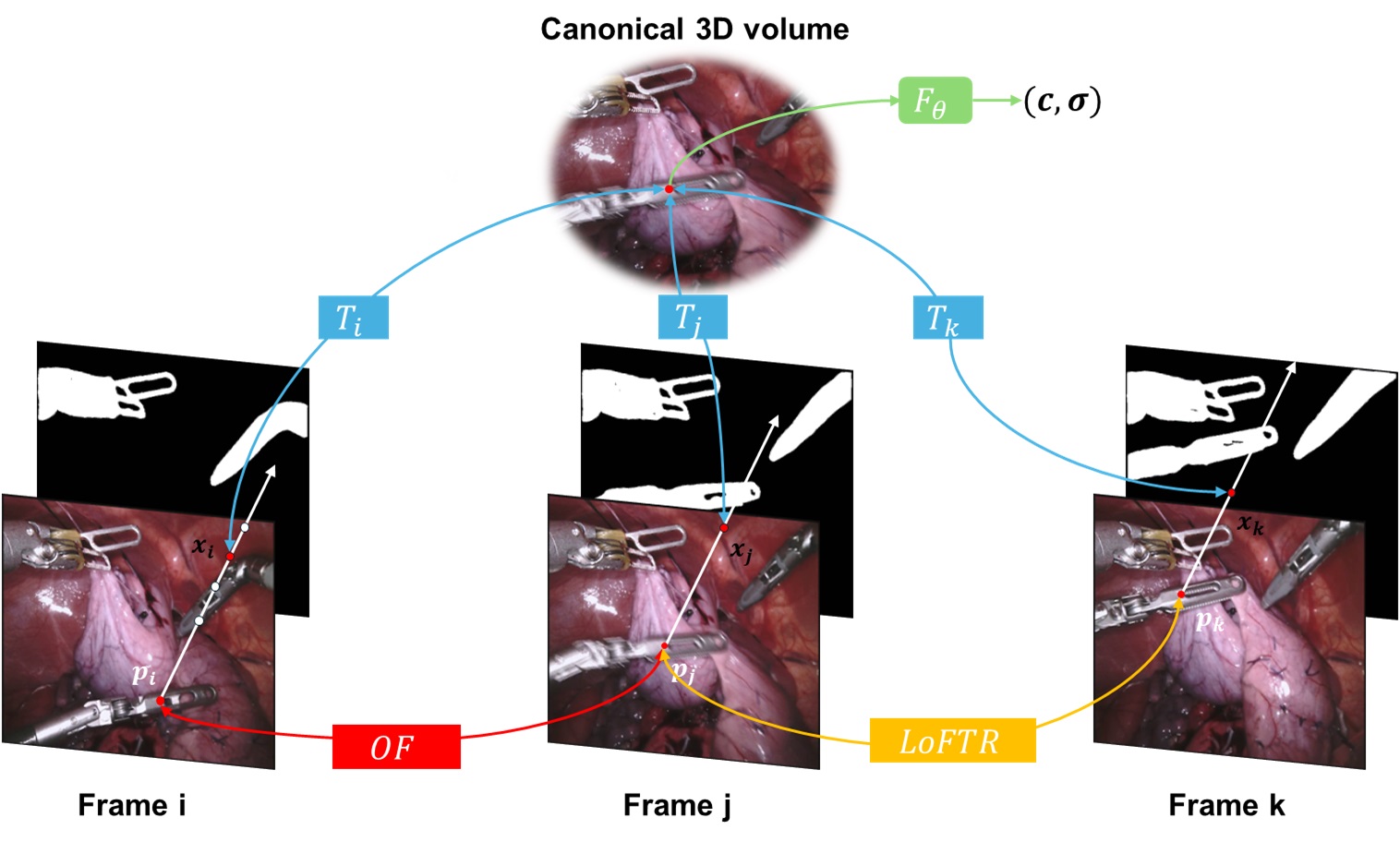}
    \vspace{0.2cm}
    \caption{Method overview: First, 2D points \(p_i\) are lifted to 3D points \(x_i\). Then, by using a bijective transformation \(T_i\), the \(x_i\) in the local frame are mapped to a canonical 3D volume as \(u\), and subsequently mapped to another local frame through an inverse bijection. A coordinate-based network \(F_\theta\) is employed to compute the corresponding color \(c\) and density \(\sigma\) of point \(u\) in the canonical volume, with the 2D positions obtained through alpha compositing. To ensure that points on the tools remain correctly mapped to the tool area, we introduce a tool mask and ARAP constraints. Additionally, since OmniMotion~\cite{wang2023tracking} is supervised by optical flow (OF), which becomes inaccurate in distant frames, we incorporate LoFTR~\cite{sun2021loftr} feature matching to enhance long-term tracking capabilities.}
    \label{fig:method}
\end{figure*}

In this work, we developed a surgical dataset with manually labeled point trajectories as the ground truth to evaluate surgical tracking algorithms. Additionally, we proposed a new method, SurgMotion, to adapt the existing TAP-based algorithm to improve its performance in surgical instrument tracking. Our approach incorporates three key loss functions: (1) a mask loss that constrains points within the instrument area, (2) an ARAP loss that enforces accurate point correspondence, and (3) a long-term loss that improves tracking accuracy across distant frames. The overview of our method can be seen in Fig. \ref{fig:method}.

\subsection{Dataset Creation}
\subsubsection{Data Sources}
Our dataset is derived from the SurgT benchmark \cite{cartucho2024surgt} and the EndoNerf dataset \cite{wang2022neural}. The videos in the SurgT benchmark originate from the Hamlyn dataset \cite{giannarou2012probabilistic}, the SCARED dataset \cite{allan2021stereo}, and the Kidney boundary dataset \cite{hattab2020kidney}. These videos cover a wide range of surgical scenes, providing a comprehensive evaluation of the algorithms, including da Vinci robotic prostatectomy, \textit{in-vivo}  porcine abdominal kidney procedures, and \textit{in-vivo} human surgeries, specifically robotic-assisted partial nephrectomy.


\subsubsection{Data Processing}
From the aforementioned datasets, we selected video clips that contain both tissues and instruments, trimming them to approximately 50 to 70 frames each. Videos with an original resolution of 1280$\times$1024 pixels were downsampled to 640$\times$512 pixels, while the remaining videos, originally at 640$\times$480 pixels, retained their original resolution

\subsubsection{Annotation Process}
To evaluate the tracking algorithms, we manually annotated the selected videos. The annotation tool was developed by adapting the SurgT-labelling tool to meet our specific requirements. Our manual annotation process basically followed the widely-used TAP-Vid-DAVIS dataset guidelines~\cite{doersch2022tap}.

For each video, we annotated 25 points per frame, consistent with the TAP-Vid-DAVIS dataset \cite{doersch2022tap}. Surgical instruments and tissues were labeled separately. Typically, each video contained two moving instruments, with 5 points annotated per instrument, totaling 10 points. The remaining 15 points were allocated for tissue annotations.

\begin{figure}[t]
\centering
\includegraphics[width=1\linewidth]{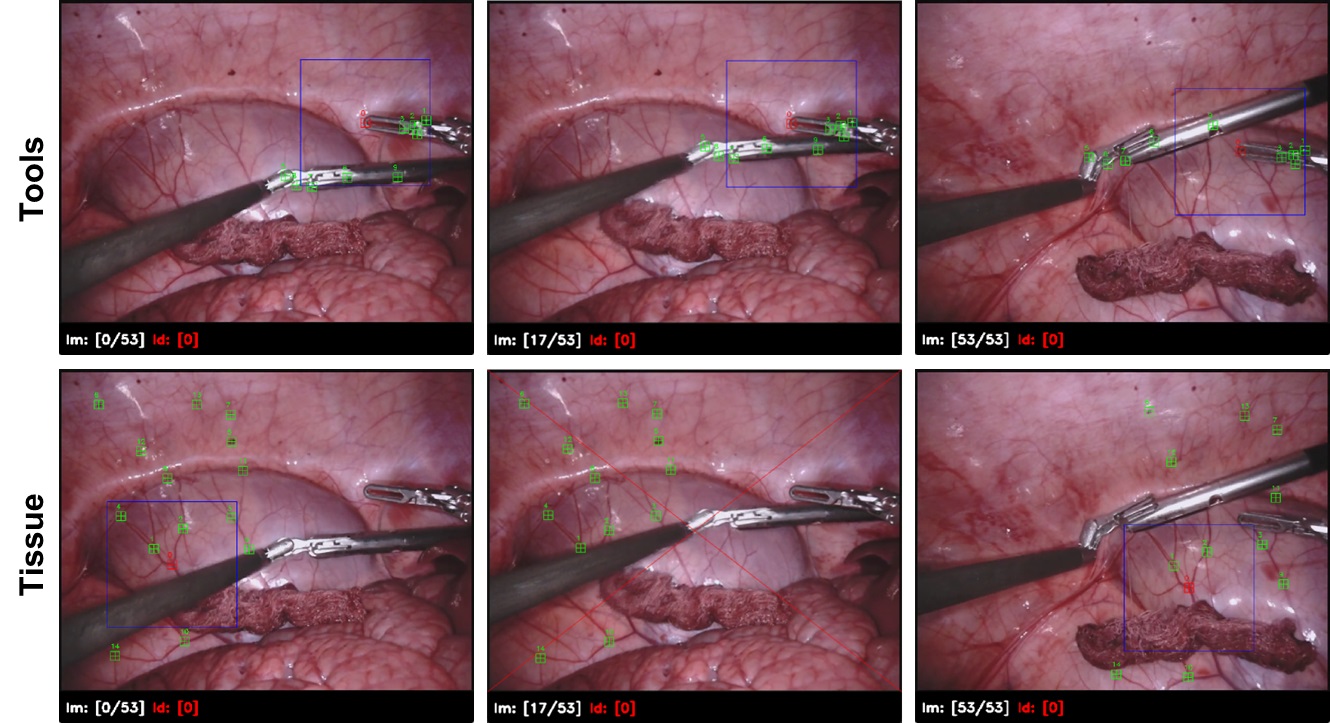}
\vspace{0.01cm}
\caption{Examples of the dataset, where tissues and surgical tools are annotated separately. The red cross indicates that this point is occluded in this frame. }
\label{fig:dataset}
\end{figure}

For surgical instruments, we selected points at the instrument tips, corners, or locations with distinctive features (e.g., screws, mechanical joints). For tissues, points were chosen in areas with textures, such as blood vessel junctions. Each selected landmark was manually tracked across all frames, with occluded points labeled as ``occluded'' and points that moved out of the frame as ``out of view''. Fig.~\ref{fig:dataset}  provides a visual example of the annotated dataset.

\subsection{Preliminaries}
OmniMotion~\cite{wang2023tracking} introduces a test-time optimization method for long-term pixel-wise tracking, even under occlusion. It represents videos as a canonical 3D volume \(G\) and tracks pixels via bijections between local frames and the canonical frame. Specifically, OmniMotion~\cite{wang2023tracking} consists of three main steps: first, the 2D point \(p_i\) is sampled along the ray orthogonal to the image plane, lifting the 2D point into a 3D point \(x_i\). Then, bijections are established between the 3D point \(x_i\) in the local frames and a point \(u\) in the canonical volume, represented as \(u = T_i(x_i)\). These bijections are parameterized as invertible neural networks Real-NVPs\cite{dinh2016density}. Using these bijections, the 3D point can be mapped from one local frame \(L_i\)  to another local frame \(L_j\):
\begin{equation}
x_j = T_j^{-1} \circ T_i (x_i)
\end{equation}
Following the NeRF \cite{mildenhall2021nerf} approach, OmniMotion~\cite{wang2023tracking} maps all 3D points \(u\in G
\) in the canonical volume to color and density using a coordinate-based MLP network \(F\), which is represented as:
\begin{equation}
(\sigma_k, c_k) = F_\theta(M_\theta(x_i^k; \psi_i))
\end{equation}
Finally, OmniMotion~\cite{wang2023tracking} projects the 3D points onto a 2D plane through volume rendering. Specifically, after applying the bijections, a set of 3D points 
\(x_i^k\) corresponding to \(p_i\) in frame \(i\) are mapped to the canonical volume \(u\), and then mapped to frame \(j\) to obtain another set of 3D points \(x_j^k\). These 3D points are aggregated via alpha compositing to produce the 3D point \(\hat{x}_j\)
\begin{equation}
\hat{x}_j = \sum_{k=1}^{K} T_k \alpha_k x_j^k, \quad \text{where} \quad T_k = \prod_{l=1}^{k-1} (1 - \alpha_l)
\end{equation}
By projecting the 3D point \(x_j\) onto the 2D plane, the 2D point \(p_j\) is obtained. The same process can also be applied to obtain the image space color \(C_j\).
The model is supervised by an optical flow loss and an RGB loss, where the predicted flow is guided by the optical flow calculated using the pre-trained RAFT model\cite{teed2020raft}.

\subsection{Proposed Method: SurgMotion}\label{sec:key_contribution}

In surgical scenarios, the motion patterns of the instruments and tissues exhibit two distinctly different characteristics. Tissues tend to undergo significant deformation but generally move slowly when manipulated with surgical instruments, making point tracking relatively straightforward. In contrast, instruments move rapidly, often resulting in motion blur in the video. Their thin and fine tips further complicate point tracking, frequently leading to substantial loss of points during the tracking process. To mitigate these challenges, we incorporate three specialized loss functions for surgical instruments to enhance tracking performance.

\textbf{Tool Mask Constraints.} 
An intuitive approach to improve instrument tracking is to ensure that points on instruments stay within their designated regions throughout the tracking process. To achieve this, we utilize existing segmentation models \cite{ma2024segment} to extract masks of the surgical instruments and introduce a mask loss to ensure that the points on the instruments consistently remain within the instrument's mask during training. This strategy helps to prevent issues where tracking points on surgical instruments are erroneously ``left behind'' on the tissue. The mask loss was defined as follows:
\begin{equation}
\mathcal{L}_{mask} = \sum_{x_i \in \Omega_M^i} \left\| M(x_i) - M(x_j) \right\|_2^2
\end{equation}
where \(\Omega_M^i\) represents the set of all pixels within the mask in frame \(i\).  \(M\) is a binary image where 1 indicates the regions occupied by the tools. 
\(x_j\) is the corresponding point of \(x_i\) in frame \(j\). The mask loss is minimized by reducing the mean squared error (MSE) between masks across frames, thereby ensuring that points belonging to instruments remain consistently within the instrument mask over time.

\textbf{As Rigid As Possible Constraints.} After applying the Tool Mask Constraints, the model effectively keeps tool points within the mask region. However, this does not guarantee that the points are correctly positioned relative to each other. Given that OmniMotion~\cite{wang2023tracking} lifts the points from the local frame to 3D during training, the 3D coordinates of these points can be accessed. With this 3D information, we can establish ARAP constraints because surgical instruments are typically rigid bodies, and the displacement of points on the same instrument are supposed to be consistent. By leveraging this property, we can enhance spatial consistency, ensuring that the points on the tools appear in their correct corresponding positions.

To enforce the ARAP constraint, the first step is to check whether the points belong to the same rigid part of the surgical tool. We employ the K-means clustering algorithm here to classify the points into different motion groups and the ARAP loss is formulated as follows:
\vspace{0.3cm}
\begin{equation}
\mathcal{L}_{arap} = \sum_{(x^k, x^p) \in \Omega_{k,p}} \left\| d(x_i^k, x_j^k) - d(x_i^p, x_j^p) \right\|_1    
\end{equation}
where \(\Omega_{k,p}\) represents the set of all pairwise points within the same rigid part, and \(d(\cdot, \cdot)\) denotes the Euclidean distance function. By minimizing the ARAP loss, the displacement between any pair of corresponding points within the same rigid cluster is enforced to be consistent.

\textbf{Sparse Feature Matching Guidance}. Even with the application of mask loss and ARAP loss, our method still encounters significant challenges in maintaining accuracy during long-term tracking. This is primarily because the method relies on optical flow for supervision and optical flow tends to result in numerous erroneous matches between distant frames. While OmniMotion~\cite{wang2023tracking} attempts to filter out these incorrect matches using cycle consistency and appearance filtering, this approach usually leads to the loss of tracking information for many points over long-term tracking, particularly those on surgical instruments. To overcome this limitation, we incorporate a feature matching algorithm to guide long-term tracking. Specifically, we use the LoFTR~\cite{sun2021loftr}, a transformer-based semi-dense feature matching method, to extract pixel-wise matches from any two frames. By integrating LoFTR~\cite{sun2021loftr} into our tracking framework, the information for points that are previously filtered out is successfully recovered. This approach not only enhances the quality of point information available for long-term tracking but also provides additional corresponding points for surgical instruments. We formulate the long-term loss as follows:

\begin{equation}
\mathcal{L}_{long} = \sum_{(p_{i,j} \in \Omega)} \left\| \left( \hat{p_j} - p_i \right) - \left( p_j - p_i \right) \right\|_1    
\end{equation}
where \(\Omega\) is the set of all pairwise points in feature matching. \(\hat{p_j}\) is the point matched to \(p_i\) by LoFTR~\cite{sun2021loftr}, and \(p_j\) is the corresponding point predicted by our model. We minimize the mean absolute error (MAE) between the predicted matching from our model and the ground truth matching generated by LoFTR~\cite{sun2021loftr}.

\section{Experiments and Results} \label{Sec:exp}

\subsection{Evaluation Metrics}
Following the TAP-Vid benchmark \cite{doersch2022tap}, we evaluate the accuracy of tracking and occlusion on our dataset and the evaluation metrics include:
\begin{itemize}
    \item   \(< \delta_{\text{avg}}^x\) represents the average position accuracy of visible points, measuring the percentage of predicted points falling within five distance thresholds: 1, 2, 4, 8, and 16 pixels from their ground truth.
\end{itemize}
\begin{itemize}
    \item Average Jaccard (AJ) assesses the joint accuracy of the predicted positions and visibility.
\[\text{AJ} = \frac{\text{True positives}}{\text{True positives} + \text{False positives} + \text{False negatives}}\]
where true positives are points within the distance thresholds of visible ground truth. False positives are points predicted as visible but are occluded or beyond thresholds, and false negatives are visible ground truth predicted as occluded or outside the thresholds.
\end{itemize}
\begin{itemize}
    \item Occlusion Accuracy (OA) measures the accuracy of visibility predictions for all points, including both visible and occluded points. 
\end{itemize}
Following the evaluation protocols of OmniMotion, we resize the images to 256$\times$256 during evaluation.

\subsection{Implementation Details}
The experiments were conducted on NVIDIA GPU3090 with PyTorch framework and the model was trained with the Adam optimizer for 100k iterations per video sequence. Each training batch includes 256 extracted point pairs from 8 pairs of frames, with 192 points sampled from optical flow and 64 points from feature matching. We assigned specific weights to the loss functions. For iterations 0 to 20k, \( w_{mask} \) and \( w_{arap} \) were set to 0, and fixed at 1 thereafter. Furthermore, \( w_{long} \), the weight for the long-term loss, was set to 0.3.

\subsection{Comparisons}
We compare our method with five SOTA TAP-based algorithms on our dataset, including RAFT~\cite{teed2020raft}, PIPs~\cite{harley2022particle}, PIPs++~\cite{zheng2023pointodyssey}, MFT~\cite{neoral2024mft}, TAPIR~\cite{doersch2023tapir}, CoTracker~\cite{karaev2023cotracker}, and OmniMotion~\cite{wang2023tracking}. These algorithms serve as baselines for evaluating the performance of our proposed method. However, tracking methods designed specifically for surgical instruments cannot track any points, making them not directly comparable.

\begin{figure*}[!h]
\centering
\includegraphics[width=0.7\linewidth]{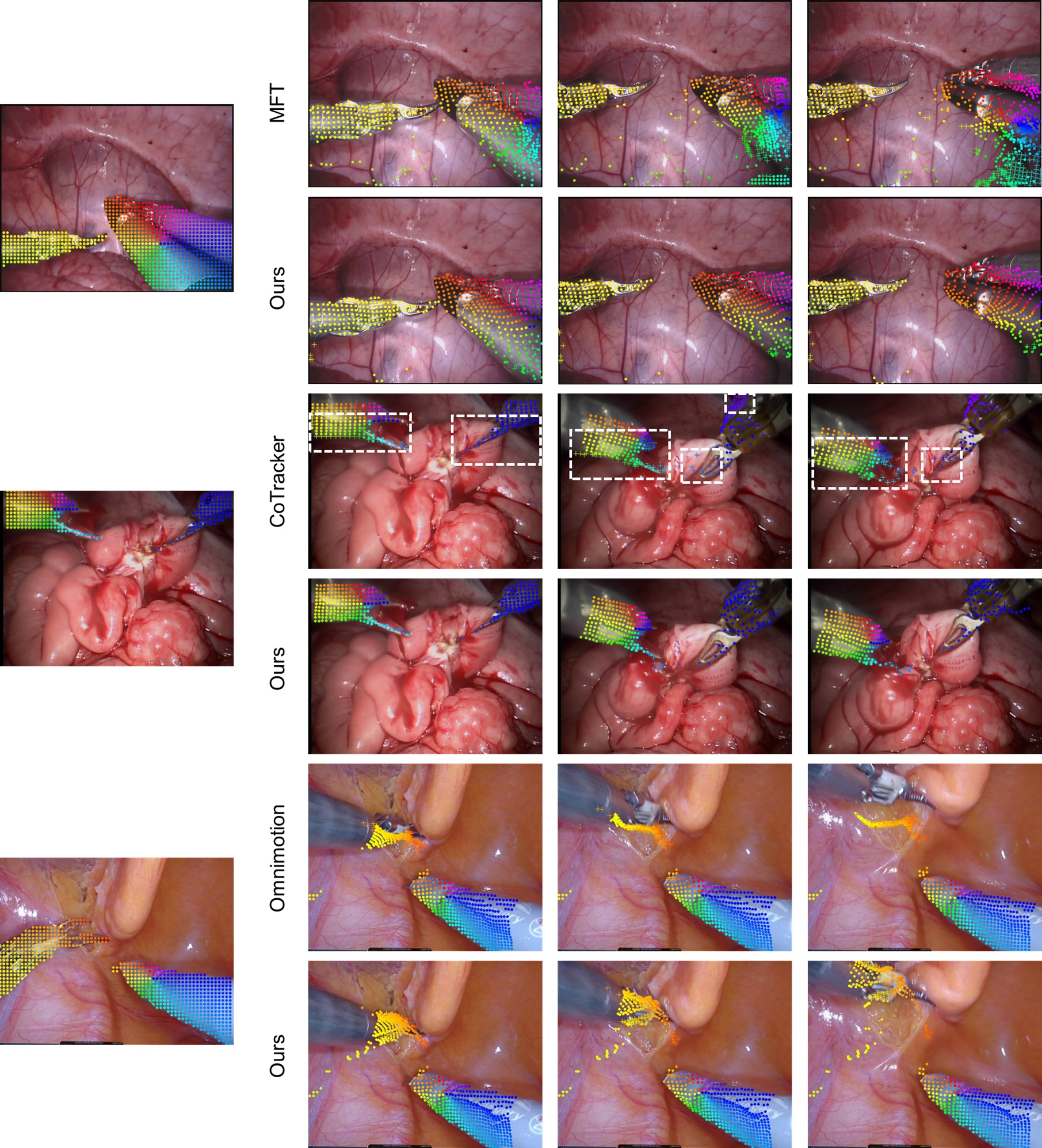}
\vspace{0.2cm}
\caption{Qualitative comparison of our method with other baselines on our dataset. The leftmost column shows the initial query points. The three columns on the right display the tracking results over time. Occluded points are marked with a cross “+” and their estimated positions are shown. Notably, the white dashed boxes highlight CoTracker's incorrect occlusion predictions, whereas our method produces accurate results in these cases. }
\label{fig:result2}
\end{figure*}

\subsubsection{Quantitative Comparisons}

We compare our method with other baselines on our dataset in Table~\ref{tab:quantitative}. In surgical instrument tracking, our method significantly outperforms RAFT~\cite{teed2020raft}, PIPs \cite{harley2022particle}, PIPs++~\cite{zheng2023pointodyssey}, MFT~\cite{neoral2024mft}, and TAPIR~\cite{doersch2023tapir}. Furthermore, it achieves SOTA in both AJ and OA metrics. Compared to our most direct competitor, OmniMotion \cite{wang2023tracking}, our method demonstrates improvements across all three metrics. Notably, it surpasses the SOTA method, CoTracker~\cite{karaev2023cotracker}, by 7.2\% in OA metric.

\begin{table}[h]
\centering
\caption{Quantitative comparison between our SurgMotion method and baselines. The optimal and suboptimal results are shown in \textbf{bold} and \underline{underlined} respectively.}
\label{tab:quantitative}
\vspace{0.2cm}
\setlength{\tabcolsep}{4pt}       
\begin{tabular}{lcccccc}
\toprule
\multirow{2}{*}{Method} & \multicolumn{3}{c}{Tools} & \multicolumn{3}{c}{Tissue} \\
\cmidrule(lr){2-4} \cmidrule(lr){5-7}
 & AJ $\uparrow$ & $<\delta^x_{avg}$ $\uparrow$ & OA $\uparrow$  & AJ $\uparrow$ & $<\delta^x_{avg}$ $\uparrow$ & OA $\uparrow$  \\
\midrule
RAFT \cite{teed2020raft}  & 49.7 & 64.8 & 91.1  & 68.5 & 84.6 & 87.6  \\
PIPs \cite{harley2022particle} & 54.1 & 69.2 & 88.2  & 58.5 & 73.8 & 88.6  \\
PIPs++ \cite{zheng2023pointodyssey}  & - & 68.3 & - & - & 83.8 & -  \\
MFT \cite{neoral2024mft} & 56.1 & 67.3 & 87.5  & 78.0 & \underline{87.7} & 94.2  \\
TAPIR \cite{doersch2023tapir} & 60.8 & 71.2 & 88.1 & 73.6 & 82.3 & 94.1  \\
CoTracker \cite{karaev2023cotracker} & \underline{62.8} & \textbf{77.1} & 85.1  & 78.3 & 86.6 & 94.1  \\
OmniMotion \cite{wang2023tracking} & 62.0 & 73.3 & \underline{91.5}  & \textbf{80.3} & \textbf{87.9} & \textbf{96.9}  \\
\midrule
SurgMotion (Ours) & \textbf{63.0} & \underline{74.2} & \textbf{92.3}  & \underline{79.8} & 87.5 & \underline{96.1}  \\
\bottomrule
\end{tabular}
\end{table}

\begin{table}[!h]
\centering
\renewcommand{\arraystretch}{1.1}  
\setlength{\tabcolsep}{10pt}       
\caption{Comparison between Baselines and our SurgMotion method on Challenging Tool Tracking Cases.}
\label{tab:challenge case}
\vspace{0.2cm}
\begin{tabular}{l c c c}  
\hline
\multirow{2}{*}{Method} & \multicolumn{3}{c}{Tools} \\
\cmidrule(lr){2-4} 
 & AJ $\uparrow$ & $<\delta^x_{avg}$ $\uparrow$ & OA $\uparrow$ \\
\midrule
RAFT \cite{teed2020raft} & 44.3 & 60.7 & 89.6 \\ 
SurgMotion & \textbf{59.2} & \textbf{71.3} & \textbf{90.8} \\  
\hline
PIPs \cite{harley2022particle} & 48.5 & 64.3 & 87.0 \\ 
SurgMotion & \textbf{57.3} & \textbf{69.4} & \textbf{90.6} \\ 
\hline
PIPs++ \cite{zheng2023pointodyssey} & - & 64.3 & - \\ 
SurgMotion & - & \textbf{71.9} & - \\ 
\hline
MFT  \cite{neoral2024mft} & 46.0 & 59.2 & 83.2 \\ 
SurgMotion & \textbf{56.9} & \textbf{69.6} & \textbf{89.9} \\ 
\hline
TAPIR \cite{doersch2023tapir} & 55.9 & 66.6 & 85.1 \\ 
SurgMotion & \textbf{59.1} & \textbf{70.7} & \textbf{91.6} \\
\hline
CoTracker \cite{karaev2023cotracker} & 49.1 & 64.8 & 79.1 \\ 
SurgMotion & \textbf{52.2} & \textbf{65.0} & \textbf{90.4} \\ 
\hline
OmniMotion \cite{wang2023tracking} & 47.3 & 61.1 & 85.7 \\ 
SurgMotion & \textbf{51.7} & \textbf{64.7} & \textbf{89.0} \\ 
\hline
\end{tabular}
\end{table}

When examining more challenging videos—defined as ``challenging cases'' where the position accuracy of visible points ($<\delta^x_{avg}$) in the baseline models is below 75\%, typically involving fast-moving instruments and motion blur that increase tracking difficulty. We compare the performance of our method with the baseline methods in these scenarios, and the results are shown in Table~\ref{tab:challenge case}. It can be seen that our method significantly outperforms all other baselines in these challenging cases, where rapidly moving instruments appear and blurred motion happens, indicating the superior capability of our method in handling such difficult scenarios and improving the overall instrument tracking performance. 

Although our method primarily targets improving tool tracking performance, without specific optimization for tissue tracking, it still maintains a high tissue tracking accuracy comparable to other baseline methods and achieves runner-up results in both AJ and OA metrics with a small margin from the SOTA. This is because tool movement is generally more salient compared to tissue movement. Consequently, our approach achieves significant advancements in challenging tool tracking scenarios while performing on a par with existing methods for tissue tracking. 


\subsubsection{Qualitative Comparisons}
We compare our method qualitatively to the baselines in Fig.~\ref{fig:result2}. Our algorithm demonstrates superior stability in instrument tracking compared to MFT and OmniMotion, especially with fast-moving surgical instruments. While baseline methods often result in a large number of lost points on the instruments, our algorithm is more effective at maintaining tracking of these points, thereby avoiding complete failure. Furthermore, compared to CoTracker, our method demonstrates improved performance in occlusion prediction. 

\subsection{Ablation Study}
We conduct ablation experiments to validate the effectiveness of our method, as shown in Table~\ref{tab:ablation}. The three proposed loss functions, $\mathcal{L}_{mask}$, $\mathcal{L}_{arap}$ are analyzed, with $\mathcal{L}_{arap}$ applied alongside the mask. We evaluated the individual effects of $\mathcal{L}_{mask}$ and $\mathcal{L}_{long}$, as well as the combined effects of $\mathcal{L}_{mask} + \mathcal{L}_{arap}$ and $\mathcal{L}_{mask} + \mathcal{L}_{long}$. The experimental results demonstrate that both $\mathcal{L}_{mask}$ and $\mathcal{L}_{long}$ individually improve performance. Furthermore, the combination of $\mathcal{L}_{mask} + \mathcal{L}_{arap}$ outperforms $\mathcal{L}_{mask}$ alone while adding $\mathcal{L}_{long}$ produces the superior overall performance compared to the baseline. 

\begin{table}[!h]
\centering
\renewcommand{\arraystretch}{1.0}  
\caption{Ablation Study of our method.}
\label{tab:ablation}
\vspace{0.2cm}
\begin{tabular}{ccc|ccc}
\toprule
\multirow{2}{*}{\textbf{\makecell{$\mathcal{L}_{mask}$}}} & \multirow{2}{*}{\textbf{\makecell{$\mathcal{L}_{arap}$}}} & \multirow{2}{*}{\textbf{\makecell{$\mathcal{L}_{long}$}}} & \multicolumn{3}{c}{\textbf{Tools}} \\
\cmidrule(lr){4-6}
& & & \textbf{AJ $\uparrow$} & \textbf{$<\delta^x_{avg}$ $\uparrow$} & \textbf{OA $\uparrow$} \\
\midrule
$-$ & $-$ & $-$ & 62.0 & 73.3 & 91.5 \\
$\checkmark$ & $-$ & $-$ & 62.1 & 73.3 & 91.5 \\
$-$ & $-$ & $\checkmark$ & 62.6 & 74.0 & 92.5 \\
$\checkmark$ & $\checkmark$ & $-$ & 62.8 & 74.2 & \textbf{92.7} \\
$\checkmark$ & $-$ & $\checkmark$ & 62.4 & 73.8 & 91.9 \\
$\checkmark$ & $\checkmark$ & $\checkmark$ & \textbf{63.0} & \textbf{74.2} & 92.3 \\
\bottomrule
\end{tabular}
\end{table}

\section{Conclusions}\label{Sec:con}
In this paper, we present a new annotated dataset pioneering the evaluation of tracking algorithms in surgical videos, and we comprehensively benchmark SOTA TAP-based algorithms on this dataset. Existing algorithms struggle to track surgical instruments accurately due to their rapid motion, which frequently leads to motion blur and occlusion in the video. To overcome these challenges, we introduce a novel method, SurgMotion, that enhances the tracking performance of surgical instruments, particularly in challenging scenarios, and achieves superior results compared to all tested algorithms.

\bibliographystyle{class/IEEEtran}
\bibliography{class/IEEEabrv,class/reference}
   
\end{document}